\documentclass[11pt]{article}
\usepackage[margin=1in]{geometry}
\usepackage{amsmath,amssymb}
\usepackage{booktabs}
\usepackage{array}
\usepackage{graphicx}
\usepackage{xcolor}
\usepackage{microtype}
\usepackage[colorlinks=true,linkcolor=blue,citecolor=blue,urlcolor=blue]{hyperref}
\newcommand{\chd}{\mathrm{CHAIR}_{d}}
\newcommand{\Gp}{G_{\mathrm{probe}}}
\newcommand{\Gs}{G_{\mathrm{sent}}}

\title{\bfseries What Does Anti-Hallucination Training Buy?\\[2mm]
\large Saying Less, Cheap Pruning, and the Rest}

\newcommand{\afl}[1]{\textit{\small #1}}
\author{%
  \textbf{Lingkai Bu}\qquad\textbf{Qian Gao}$^{*}$\\[3pt]
  \afl{Key Laboratory of Computing Power Network and Information Security,}\\
  \afl{Ministry of Education}\\
  \afl{Shandong Computer Science Center (National Supercomputer Center in Jinan)}\\
  \afl{Qilu University of Technology (Shandong Academy of Sciences), Jinan, China}\\[3pt]
  \texttt{\small 10431250007@stu.qlu.edu.cn}\quad\texttt{\small gq@qlu.edu.cn}\\[2pt]
  \afl{$^{*}$Corresponding author.}
}
\date{\today}

\begin{document}
\maketitle

\begin{abstract}
Preference training is the standard recipe for reducing object hallucination in vision--language models,
and each new method reports a lower CHAIR score. We ask what that reduction consists of. Every trained
system is compared with its base model's caption truncated to the same object recall, which removes the
part of the gain that comes from saying less; the remainder is compared with what a deliberately weak,
frozen, eleven-feature post-hoc probe fitted on 500 annotated images can reproduce. All criteria were
specified in the analysis code and frozen before the outputs they judge existed.
For eight public methods on LLaVA-1.5-7B, the probe reproduces at least two thirds of what training
removes beyond saying less for 7 of 8 methods on COCO, 8 of 8 on AMBER and 6 of 8 on Object HalBench,
with RLAIF-V the consistent exception; a later public method released for both LLaVA-1.5 and
Qwen2.5-VL (SENTINEL) meets the same bar on both.
On our own DPO checkpoints the accounting holds on Qwen2.5-VL but fails on InternVL2.5 on COCO
(2 of 4 training designs).
At matched recall, the base model plus the probe beats truncation for 8 of 9 LLaVA-1.5 systems on COCO
and 9 of 9 on AMBER, on the two newer backbones, and when realized as actual sentence deletion.
The same measure prices training signals: DPO on the model's own samples labelled by the probe or by an
object-level CLIP score recovers 0.70 and 0.87 of the beyond-brevity gain of ground-truth-labelled DPO
on Qwen2.5-VL (the probe 0.73 on InternVL2.5), both beat length-matched random labels, and CLIP labels
buy more than probe labels on COCO.
Human judgements show that the sentences the probe deletes are wrong more often than those truncation
would delete; a human check of the CHAIR evaluator's precision falls below our pre-set bar, which we
report as a limitation.
\end{abstract}

\section{Introduction}
\label{sec:intro}
A detailed caption produced by a large vision--language model (LVLM) routinely names objects that are
not in the image. The standard measure of this is CHAIR \cite{rohrbach2018}, and preference-based training
has become the dominant recipe for lowering it: eight methods with public weights for LLaVA-1.5-7B
\cite{llava15} alone report substantial reductions
\cite{rlaifv,hadpo,opadpo,eos,halva,povid,csr,chairdpo}.

There is a well-known confound: a model that says less hallucinates less. The concern has been raised
before --- for decoding-time methods \cite{mirage,playingsafe}, and by training-based work that notes a trade-off
with coverage, length or expressiveness \cite{eos,ovip} --- but it has been addressed by reporting hallucination
and informativeness side by side or by correlating them. For the training-based family there is no
counterfactual: nobody asks what the \emph{same} model would have achieved by simply saying less.

\begin{figure}[t]
\centering
\includegraphics[width=\textwidth]{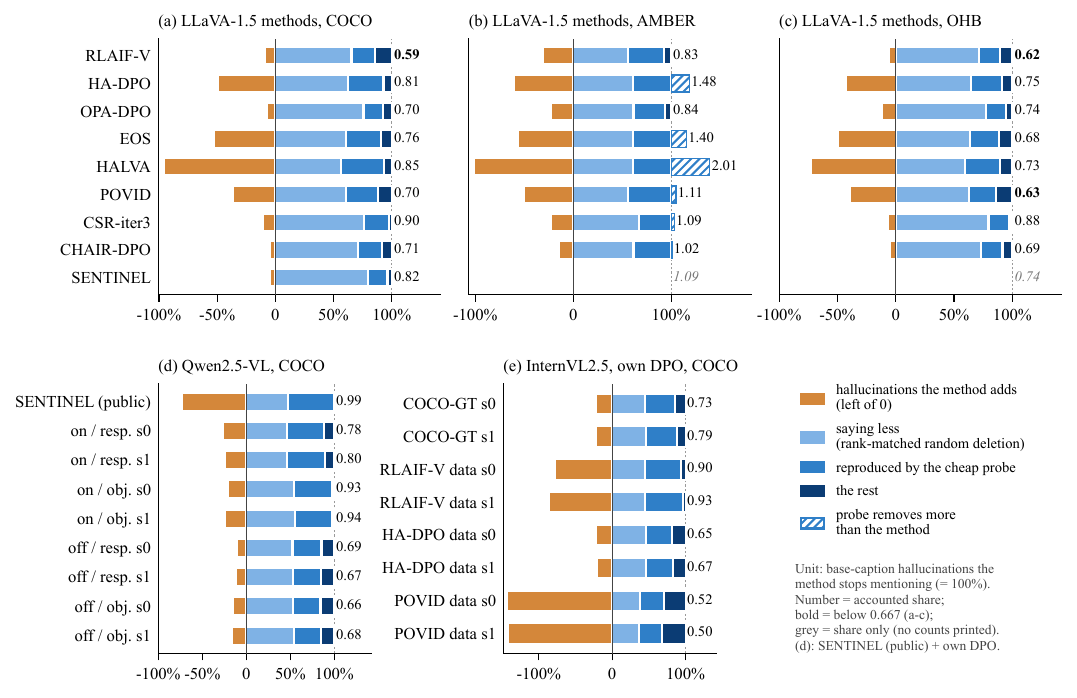}
\caption{\textbf{What training removes, decomposed.} Unit: hallucinated objects of the base model's
captions that a trained system no longer mentions on held-out images (= 100\%). Light blue: removed by
position-matched random deletion (saying less); mid blue: additionally reproduced by the frozen cheap
probe at the same per-rank counts; dark blue: the rest. Hatched: the probe removes more than the method.
Orange, left of zero: hallucinated objects the system newly introduces, on the same scale --- the
decomposition is of removals, not of the net CHAIR change. Numbers: accounted share
(Section~\ref{sec:accounting}); bold: below the pre-set bar of 0.667. Grey italic: share only (the frozen
output prints no per-rank counts for that cell). (a)--(c): eight public methods and SENTINEL on
LLaVA-1.5-7B; (d) SENTINEL and our DPO checkpoints on Qwen2.5-VL-7B; (e) our DPO checkpoints on
InternVL2.5-8B.}
\label{fig:decomp}
\end{figure}

We build that counterfactual and use it to decompose what training buys. Every trained system is
compared with its base model's caption truncated to the same object recall, and, at the level of
individual objects, with a random deletion that removes exactly as many objects at every position in the
caption as the trained method did. What training removes is thereby split into three parts
(Figure~\ref{fig:decomp}): \textbf{(1) saying less} --- what position-matched random deletion already
removes; \textbf{(2) cheap pruning} --- the part of the remainder that a frozen, cheap post-hoc probe
reproduces; and \textbf{(3) the rest}.

The probe is deliberately weak. Post-hoc hallucination detectors that are far stronger exist
\cite{lure,metatoken,haloprobe}; ours is an eleven-feature logistic regression over the finished caption's
surface statistics, COCO class priors and the model's own token log-probabilities, fitted once on 500
annotated images and frozen. A probe that ranks hallucinations better could only raise the share it
reproduces, so the shares we report are conservative.

Every headline number comes from a criterion that was written into the analysis code, together with its
decision rule, and frozen (by md5) before the outputs it judges existed. This is an internal freeze, not an
external registration; Appendix~\ref{app:prereg} lists all 32 primary criteria and their outcomes,
including 5 failures, and states for each how the freeze is documented.

Our contributions:
\begin{itemize}
\item \textbf{A matched-recall decomposition protocol} for training-based anti-hallucination methods, with
image-level bootstrap intervals and pre-specified criteria (Section~\ref{sec:protocol}).

\item \textbf{An accounting.} For 7 of 8 public LLaVA-1.5 methods on COCO, 8 of 8 on AMBER and 6 of 8 on
Object HalBench, the frozen probe reproduces at least two thirds of what training removes beyond saying
less; RLAIF-V is the consistent exception. A recent public method released for both LLaVA-1.5 and
Qwen2.5-VL, SENTINEL, meets the same bar on both. On our own DPO checkpoints the accounting holds on
Qwen2.5-VL and \emph{fails} on InternVL2.5 on COCO (Section~\ref{sec:accounting}).

\item \textbf{A strong baseline.} At matched recall, the base model plus the probe beats truncation for
8 of 9 LLaVA-1.5 systems on COCO and 9 of 9 on AMBER, on Qwen2.5-VL and on all eight InternVL2.5
checkpoints, and when the edit is realized as actual sentence deletion (Section~\ref{sec:baseline}).
Human judgements confirm that the sentences it deletes are wrong more often than those truncation would
delete (Section~\ref{sec:human}).

\item \textbf{Cheap labels suffice for preference training --- and a way to price them.} With the same
measure, the recovery ratio of a labelling source is the share of ground-truth-labelled DPO's
beyond-brevity gain it recovers, checked against length-matched random labels. Two cheap object-level
labellers calibrated on the same 500 images, our probe and a CLIP similarity score, recover 0.70 and 0.87
on Qwen2.5-VL (the probe 0.73 on InternVL2.5) and both beat matched random labels; CLIP labels buy more
than probe labels on COCO (Section~\ref{sec:labels}).
\end{itemize}

Taken together: most of what this family of training removes beyond saying less is both
\emph{reproducible} by cheap post-hoc signals and \emph{trainable} from them. The claim is
methodological --- a reusable, falsifiable protocol for auditing claimed gains and for valuing training
signals --- not that cheap probes can replace training.

\section{Related work}
\label{sec:related}
\paragraph{Training-based mitigation.} HA-DPO \cite{hadpo} frames hallucination as preference selection over
paired hallucinatory and corrected responses. RLAIF-V \cite{rlaifv} builds open-source AI feedback for
preference learning. OPA-DPO \cite{opadpo} argues that on-policy construction of the preference data is
what makes DPO work for this problem. HALVA \cite{halva} performs data-augmented phrase-level alignment,
and the EOS-selective objective \cite{eos} teaches the model when to stop. POVID \cite{povid} generates
dispreferred responses by injecting hallucinations with a stronger model and by distorting the image;
CSR \cite{csr} iterates calibrated self-rewarding; CHAIR-DPO \cite{chairdpo} ranks the model's own samples
with CHAIR computed from an off-the-shelf detector. All eight release weights for LLaVA-1.5-7B
\cite{llava15}, which is why they are the systems audited here. SENTINEL \cite{sentinel} builds
sentence-level in-domain preference data, cross-checking objects with two open-vocabulary detectors, and
releases weights for both LLaVA-1.5-7B and Qwen2.5-VL-7B \cite{qwen25vl}; for InternVL2.5 \cite{internvl25}
the only released preference-tuned weights we found are those of MPO \cite{mpo}, a preference-optimization
method aimed at multimodal reasoning rather than at hallucination, which we do not evaluate.

\paragraph{Post-hoc detection and revision.} LURE \cite{lure} revises a description using co-occurrence,
uncertainty and object position --- nearly the feature groups of our probe. MetaToken \cite{metatoken} is a
lightweight token-level detector. HaloProbe \cite{haloprobe} combines attention and decoder statistics
with external features (position, repetition, first occurrence) and mitigates by hallucination-aware beam
search or post-hoc editing. Our probe is deliberately a weak member of this family; the point of the paper
is the control it is measured under and what it is compared with, not the feature set.

\paragraph{Auditing reported gains.} Yin et al.~\cite{mirage} attribute the gains of contrastive decoding to
its adaptive plausibility constraint. Fazli et al.~\cite{playingsafe} report that faithfulness gains of six
decoding-time methods travel with losses of informativeness and argue for a three-way trade-off view;
their scope explicitly excludes training-based methods, and their evidence is correlational. Our work is
the training-side counterpart, with a constructed matched counterfactual and bootstrap intervals. Feng et
al.~\cite{feng2024} find that CHAIR's object-level judgements are frequently wrong; our own human check of
the evaluator (Section~\ref{sec:human}) points the same way, and we report it as a limitation.

\paragraph{Cheap labels for preference training.} Several methods already train DPO on labels from cheap
signals: CLIP-DPO \cite{clipdpo} ranks sampled responses by CLIP image--text similarity, CHAIR-DPO
\cite{chairdpo} by CHAIR computed with an off-the-shelf object detector, SENTINEL \cite{sentinel} by
cross-checking two open-vocabulary detectors (GroundingDINO and YOLO-World), and HSA-DPO \cite{hsadpo} with a
hallucination detector trained on sentence-level annotations produced by proprietary models. We do not claim that cheap labelling is new. Our CLIP
labeller is an object-level adaptation inspired by CLIP-DPO, not a reproduction of it. What we add is a
way to \emph{price} a labelling source under the matched-recall control --- the share of
ground-truth-labelled DPO's beyond-brevity gain it recovers, against length-matched random labels --- and
the observation that the same cheap signals that account for most of what training removes can also
supply the labels.

\section{Protocol}
\label{sec:protocol}
\subsection{Systems and data}

\textbf{Backbones and systems.} LLaVA-1.5-7B \cite{llava15} with the eight public methods above and
SENTINEL; Qwen2.5-VL-7B-Instruct \cite{qwen25vl} with SENTINEL and eight LoRA-DPO checkpoints we trained
(Section~\ref{sec:acc-own}); InternVL2.5-8B \cite{internvl25} with eight LoRA-DPO checkpoints we trained.
Every system is run under one protocol: the prompt ``\texttt{Please describe this image in detail.}'' and
greedy decoding.

\textbf{Data.} 2000 COCO \texttt{val2014} images in three disjoint batches: $A$ (500), $B$ (500) and
$C$ (1000). Batch $A$ is the \emph{only} data used to fit anything. Batches $B{+}C$ (1500 images) are the
held-out evaluation set and are never refitted, recalibrated or threshold-tuned on. Two further caption
benchmarks repeat the analyses: AMBER \cite{amber} (1004 non-COCO images, generative task, with its official
annotation, vocabulary and evaluator) and Object HalBench \cite{rohrbach2018,rlhfv} (300 COCO images with different
prompts; 295 after removing images that overlap batch $A$). POPE \cite{pope} is used for a separate,
discriminative question (Section~\ref{sec:rest}).

\subsection{Metrics}

We evaluate at the level of \emph{distinct} objects, which removes the sensitivity of mention-counted
CHAIR to how often a caption repeats an object. For a set of captions, let $\mathcal{O}_i$ be the
distinct COCO objects named in image $i$'s caption and $\mathcal{G}_i$ its ground-truth object set:
\begin{equation}
\chd \;=\; 100\cdot\frac{\sum_i |\mathcal{O}_i \setminus \mathcal{G}_i|}{\sum_i |\mathcal{O}_i|},
\qquad
\mathrm{Recall} \;=\; 100\cdot\frac{\sum_i |\mathcal{O}_i \cap \mathcal{G}_i|}{\sum_i |\mathcal{G}_i|} .
\end{equation}
Object extraction uses the standard CHAIR synonym lists \cite{rohrbach2018} with two rule-based
corrections (``orange'' as a colour, ``bear'' as a verb). On AMBER the official evaluator's object labels
take the place of the COCO annotation, with its hallucination rate and coverage in place of $\chd$ and
recall. Confidence intervals are 1000-draw image-level bootstraps (seed 0) throughout.

\subsection{The say-less control}
\label{sec:control}

The zero-cost way to lower $\chd$ is to say less. For each system, \textsc{Trunc} keeps only the first
$k$ distinct objects of its own caption, tracing a curve in the (recall, $\chd$) plane. Any pruner is then
read off \emph{at the same recall}:
\begin{equation}
\Gp(r) \;=\; \chd\big(\textsc{Trunc}\text{ at recall } r\big) \;-\; \chd\big(\textsc{Probe}\text{ at recall } r\big),
\label{eq:g}
\end{equation}
in points; $\Gp > 0$ means that the pruner does better than being shorter. The pre-specified operating
point is $r = 80\%$ of each system's own full-length recall.

\subsection{Position, held fixed, and the accounted share}
\label{sec:share}

Hallucination rises steeply with an object's position in the caption: on the held-out set the base
LLaVA-1.5 model's hallucination rate by first-mention rank is $2.5$, $18.1$, $36.0$, $47.4$, $55.7$ and
$57.2\%$ for ranks $1$--$5$ and $6{+}$. 
A comparison that does not hold position fixed mostly measures position. For each trained method we
therefore count, per rank, how many objects of the base caption it no longer mentions, and construct two
pruners on the base caption that drop \emph{exactly as many objects at every rank}: a random one, which by
construction carries no information beyond position, and the probe, which drops the highest-scoring
objects. With $R_{\mathrm{rand}}$, $R_{\mathrm{probe}}$ and $R_{\mathrm{method}}$ the numbers of
base-caption hallucinations removed by the random pruner, by the probe and by the method,
\begin{equation}
\mathrm{share} \;=\; \frac{R_{\mathrm{probe}} - R_{\mathrm{rand}}}{R_{\mathrm{method}} - R_{\mathrm{rand}}}
\end{equation}
is the fraction of the method's \emph{beyond-position} removal that the probe reproduces. It can exceed 1
when, at the same per-rank counts, the probe removes more base hallucinations than the method itself.

\subsection{Pre-specification}
\label{sec:prespec}

Each analysis is a script whose docstring states its criteria, their thresholds and the decision rule
(including when the line of work is to be abandoned) before the outputs it judges are generated; the
script and its inputs' manifest are then frozen by md5. Each new analysis script must first reproduce
previously frozen numbers (anchor gates) before any new number it prints is read; an anchor failure stops
the analysis. No result is rescued by changing the operating point, the feature set, the bootstrap or the
systems. The freeze is internal to the project, not an external registration, and the documentation
differs between analyses: for most, the md5 freeze demonstrably predates the outputs; for the first
LLaVA analyses only the scripts' own dated statements document the order; the AMBER analysis was
re-frozen once after a label-leakage bug was found in its first version, and both versions are reported
(Appendix~\ref{app:prereg}). Analyses added after seeing results are labelled \emph{post hoc}.

\section{The cheap probe}
\label{sec:probe}
The probe assigns each distinct object named in a caption a probability that it is hallucinated, from
eleven features in three groups.
\begin{description}
\item[Surface (5)] \texttt{tok\_frac}, the first mention's token offset normalized by caption length;
\texttt{rank}, its bucketed rank among distinct objects ($0..5^{+}$); \texttt{n\_ment}, how many times the
caption mentions the object; \texttt{n\_objs}, how many distinct objects the caption names;
\texttt{hedge}, whether a hedging or backgrounding marker (``possibly'', ``in the background'',
``another'', \ldots) occurs within $\pm 5$ tokens.
\item[Priors (3)] \texttt{cls\_prior}, the smoothed hallucination rate of the object's COCO class;
\texttt{cooc}, the log-odds that the class is present given the classes named earlier in the same caption,
from COCO co-occurrence; \texttt{gt\_freq}, the class's log frequency. All three are estimated on batch $A$
only.
\item[Confidence (3)] \texttt{c\_first}, \texttt{c\_min}, \texttt{c\_mean}: the log-probability of the
first, the minimum and the mean over the caption tokens of the object word, obtained by teacher-forcing
the stored caption through the model that produced it (for greedy captions, the confidence the model had
when generating them).
\end{description}
A standardizing scaler and a logistic regression ($C = 1$) are fitted on batch $A$ and frozen. A
\textsc{text-only} variant (surface + priors) needs no forward pass at all.

\textbf{Which probe is used where.} For the accounted share, the probe of the \emph{base} model is applied to
the base captions. For the strong baseline, each system is scored by its own probe fitted on its own
batch-$A$ captions. On AMBER, which has no COCO annotation, the probe is fitted on AMBER's own labels by
2-fold cross-fitting (502 images per fold); a zero-shot variant that uses the COCO-fitted probe and no
AMBER label is reported separately. On Object HalBench the COCO-fitted probe is applied unchanged.

\textbf{Pruning rule.} Objects are removed in decreasing probe score under a \emph{global} threshold, so the
probe may spend its deletions unevenly across images, whereas \textsc{Trunc} keeps the same $k$ in every
image. This asymmetry is part of the protocol and is held fixed everywhere.

\textbf{Exploratory motivation.} On batch $A$ the text-only probe attains a pooled AUROC of $0.845$ for
predicting the base model's own hallucinations, and $0.866$ with confidence added; the model's own
confidence alone attains $0.707$ and position alone $0.680$. Within a rank bucket the text-only AUROC is
$0.75$, $0.79$, $0.74$, $0.67$, $0.62$ and $0.61$ for ranks $1$--$6^{+}$, so the signal is not merely
position. These numbers motivated the criteria below and are not themselves claims.

\section{Accounting: how much of what training removes is cheap?}
\label{sec:accounting}
We report the accounted share of Section~\ref{sec:share}, with the base model's frozen probe. The pre-set
bar is $\mathrm{share} \ge 0.667$. It was chosen with reference to an earlier internal result
in which a static confidence threshold reproduced $46$--$57\%$ and was recorded as a failure, so that
anything near $0.6$ cannot pass. On LLaVA-1.5 the criterion requires at least 6 of the 8 methods to reach
the bar; on our own checkpoints, at least 3 of 4 training designs (mean over two seeds).

\subsection{Eight public methods on LLaVA-1.5}

\begin{table}[t]
\centering
\small
\begin{tabular}{lrrrcccc}
\toprule
& \multicolumn{4}{c}{COCO held-out (1500 images)} & & AMBER & OHB \\
\cmidrule(lr){2-5}
method & $R_{\mathrm{rand}}$ & $R_{\mathrm{probe}}$ & $R_{\mathrm{method}}$ & share & & share & share \\
\midrule
RLAIF-V    & 780 & 1021 & 1192 & \textbf{0.59} & & 0.83 & \textbf{0.62} \\
HA-DPO     & 461 &  686 &  738 & 0.81 & & 1.48 & 0.75 \\
OPA-DPO    & 933 & 1142 & 1232 & 0.70 & & 0.84 & 0.74 \\
EOS        & 452 &  671 &  741 & 0.76 & & 1.40 & 0.68 \\
HALVA      & 274 &  457 &  489 & 0.85 & & 2.01 & 0.73 \\
POVID      & 517 &  751 &  850 & 0.70 & & 1.11 & \textbf{0.63} \\
CSR-iter3  & 820 & 1050 & 1075 & 0.90 & & 1.09 & 0.88 \\
CHAIR-DPO  & 817 & 1050 & 1144 & 0.71 & & 1.02 & 0.69 \\
\midrule
criterion  & \multicolumn{4}{c}{7/8 $\ge 0.667$: PASS} & & 8/8: PASS & 6/8: PASS \\
\midrule
SENTINEL$^\S$ & 949 & 1150 & 1193 & 0.82 & & 1.09 & --- \\
\bottomrule
\end{tabular}
\caption{\textbf{Accounting on LLaVA-1.5-7B.} Hallucinated objects of the base captions removed by the
position-matched random pruner, by the frozen probe and by the method, at exactly matched per-rank
counts; share $= (R_{\mathrm{probe}}-R_{\mathrm{rand}})/(R_{\mathrm{method}}-R_{\mathrm{rand}})$.
Bold: below 0.667. AMBER: 1004 non-COCO images, probe cross-fitted on AMBER labels; Object HalBench (OHB):
295 COCO images with different prompts, COCO-fitted probe applied unchanged. $^\S$SENTINEL was evaluated
later under its own criteria (S1, S5), judged on its own and not part of the 8-method verdicts.}
\label{tab:acc}
\end{table}

On COCO, 7 of the 8 methods have shares between 0.70 and 0.90 (Table~\ref{tab:acc}): for them, at least
seven tenths of what they remove beyond position is removed by the base model plus a frozen
eleven-feature probe at the same per-rank counts. The exception is RLAIF-V at 0.59. A text-only probe,
which needs no forward pass, gives shares lower by $0.06$--$0.26$, so the model's confidence contributes
but is not the main source. The same protocol, again with pre-specified criteria, passes on AMBER
(8/8; five methods exceed 1, e.g.\ for HALVA the probe removes 258 base hallucinations at the per-rank
counts where HALVA itself removes 185) and on Object HalBench (6/8; RLAIF-V 0.62 and POVID 0.63 fall
below). With the COCO-fitted probe and no AMBER label at all, the AMBER accounting still passes (6/8).
RLAIF-V has the lowest share on all three benchmarks. We report this as an observation only: a
pre-specified test of the reading that RLAIF-V learns information the cheap signals lack did not
replicate on AMBER (Section~\ref{sec:rest}).

\subsection{A public method on a newer backbone: SENTINEL}

The eight methods above are all built on LLaVA-1.5. To address the objection that public methods exist
only for an old model, we searched for public anti-hallucination weights for Qwen2.5-VL and InternVL2.5.
The only object-hallucination method with released Qwen2.5-VL-7B weights we found is SENTINEL
\cite{sentinel}, which also released LLaVA-1.5-7B weights. Its criteria were frozen before its adapters
were downloaded, with four primary criteria judged one by one. On LLaVA-1.5 (S1) its share is 0.82
($R_{\mathrm{rand}}$ 949, $R_{\mathrm{probe}}$ 1150, $R_{\mathrm{method}}$ 1193), so with SENTINEL as a
ninth method 8 of 9 reach the bar, RLAIF-V still the only one below. On Qwen2.5-VL (S3) its share is 0.99
($R_{\mathrm{rand}}$ 118, $R_{\mathrm{probe}}$ 246, $R_{\mathrm{method}}$ 247), but its beyond-position
removal is only 129 objects (the pre-set minimum for evaluability was 30), so this number is small-sample.
Excluding the held-out images that also occur in SENTINEL's training data (15 for LLaVA, 8 for Qwen)
leaves the shares at 0.83 and 0.98 (descriptive). For SENTINEL, then, the accounting holds on both
backbones; one public method on Qwen2.5-VL is not evidence about public methods on new backbones in
general.

\subsection{Our own checkpoints on two newer backbones}
\label{sec:acc-own}

Because public methods on newer backbones are scarce, we trained eight LoRA-DPO checkpoints on each of
Qwen2.5-VL-7B and InternVL2.5-8B (2527 preference pairs each, COCO \texttt{train2014} images or public
preference data, identical recipe).

\textbf{Qwen2.5-VL-7B} ($2\times2\times2$ seeds): preference data either from Qwen's own samples (on-policy)
or from LLaVA-1.5's samples on the same images (off-policy), at response granularity (best vs.\ worst
sample) or object granularity (worst sample vs.\ itself without its hallucinated sentences). The primary
question of this experiment was whether on-policy training removes what the probe cannot reproduce; its
answer is negative (Section~\ref{sec:rest}). Accounting is a \emph{secondary} criterion here: the
two-seed mean shares are 0.79 (on-policy, response), 0.94 (on-policy, object), 0.68 (off-policy, response)
and 0.67 (off-policy, object; seeds 0.66 and 0.68, at the bar), 4 of 4 reaching 0.667.

\textbf{InternVL2.5-8B} (4 data sources $\times$ 2 seeds): InternVL's own samples labelled with the COCO
annotation (GT), and the public RLAIF-V, HA-DPO and POVID preference data, each sampled to 2527 pairs after
removing every pair whose image is one of our evaluation images. Accounting is the \emph{primary}
criterion. The two-seed mean shares are 0.76 (GT), 0.92 (RLAIF-V data), 0.66 (HA-DPO data) and 0.51
(POVID data): 2 of 4 designs reach the bar where 3 were required, so \textbf{the accounting fails on
InternVL2.5 on COCO}. A descriptive repetition on AMBER gives 1.13, 1.25, 0.97 and 0.94 (4 of 4), so the
failure is specific to COCO. Both failing designs use preference data written by other models (GPT-4
rewrites for HA-DPO, injected hallucinations for POVID); we report this as an observation, since the
causal test we pre-specified (style-matched rewriting) did not pass its pilot and was stopped.

\subsection{Removed versus added}

Figure~\ref{fig:decomp} decomposes removals, not the net change in CHAIR: trained systems also name
hallucinated objects the base caption did not. In a post-hoc count on COCO, the ratio of added to removed
hallucinated objects is 0.96 for HALVA (489 removed, 467 added: net 22), 0.53 for EOS, 0.49 for HA-DPO,
0.36 for POVID, and at most 0.11 for the other four. For SENTINEL it is 0.05 on LLaVA-1.5 and 0.73 on
Qwen2.5-VL (247 removed, 181 added): on Qwen it changes little, and most of what it removes is offset by
what it adds.

In sum, the accounting holds for the public methods on LLaVA-1.5 on three caption benchmarks, for
SENTINEL on two backbones, and for our checkpoints on Qwen2.5-VL; it does not hold for our checkpoints on
InternVL2.5 on COCO.

\section{A strong baseline: the probe beats saying less}
\label{sec:baseline}
The accounting asks how much of what training does a probe can reproduce. The strong baseline asks the
converse: applied on top of any system, including a trained one, does probe-guided pruning do better
than that system simply saying less? We report $\Gp$ (Eq.~\ref{eq:g}) at 80\% of each system's own full
recall, each system scored by its own probe fitted on its own batch-$A$ captions.

\begin{table}[t]
\centering
\small
\setlength{\tabcolsep}{3pt}
\begin{tabular}{lcccc}
\toprule
system & full recall & COCO $\Gp$ & AMBER $\Gp$ & OHB $\Gp$ \\
\midrule
LLaVA-1.5-7B & 79.7 & $+4.37$ $[+3.28,+5.52]$ & $+2.85$ $[+2.13,+3.60]$ & $+4.12$ $[+1.73,+6.80]$ \\
RLAIF-V      & 65.9 & $+1.63$ $[+0.88,+2.35]$ & $+1.97$ $[+1.38,+2.57]$ & $+2.09$ $[-0.16,+4.08]$ \\
HA-DPO       & 75.6 & $+3.61$ $[+2.48,+4.64]$ & $+2.35$ $[+1.53,+3.22]$ & $+4.71$ $[+2.44,+6.60]$ \\
OPA-DPO      & 56.1 & $+0.85$ $[+0.09,+1.64]$ & $+0.98$ $[+0.50,+1.53]$ & $+2.05$ $[-0.34,+4.20]$ \\
EOS          & 76.9 & $+3.13$ $[+2.23,+4.01]$ & $+1.83$ $[+1.21,+2.49]$ & $+3.77$ $[+1.59,+6.20]$ \\
HALVA        & 79.9 & $+4.57$ $[+3.55,+5.76]$ & $+2.43$ $[+1.63,+3.17]$ & $+5.40$ $[+3.19,+7.95]$ \\
POVID        & 75.7 & $+4.52$ $[+3.52,+5.41]$ & $+2.83$ $[+2.03,+3.57]$ & $+5.06$ $[+2.87,+7.06]$ \\
CSR-iter3    & 63.0 & $+2.70$ $[+1.86,+3.78]$ & $+1.87$ $[+1.28,+2.55]$ & $+1.21$ $[-0.81,+3.15]$ \\
CHAIR-DPO    & 63.5 & $+0.53$ $[-0.24,+1.37]$ & $+1.16$ $[+0.62,+1.76]$ & $+1.43$ $[-0.18,+3.38]$ \\
\midrule
criterion & & 8/9 CI $>0$: PASS & 9/9: PASS & mean $+3.31$ $[+2.16,+4.52]$: PASS \\
\midrule
SENTINEL$^\S$ & 57.0 & $+1.86$ $[+1.07,+2.64]$ & $+1.85$ $[+1.29,+2.38]$ & $+1.45$ $[+0.11,+2.98]^\dagger$ \\
\bottomrule
\end{tabular}
\caption{\textbf{Strong baseline on LLaVA-1.5-7B.} $\Gp$ at 80\% of each system's own full recall, in points;
95\% image-bootstrap intervals; positive means probe-guided pruning beats truncating the same caption to
the same recall. OHB criterion: $\Gp > 0$ for at least 8 of 9 systems and the mean's interval above zero;
single-system intervals containing zero are listed as they are. $^\S$Evaluated later and judged on its
own (S2, S5); its full recall is far below the base model's (57.0 vs.\ 79.7). $^\dagger$Descriptive.}
\label{tab:baseline}
\end{table}

\textbf{LLaVA-1.5.} On COCO the probe beats truncation for 8 of 9 systems (Table~\ref{tab:baseline}),
including on top of each trained method's own output --- training did not absorb the cheap signal. The one
interval that contains zero is CHAIR-DPO's ($+0.53$ $[-0.24,+1.37]$), which we describe as not
significant rather than as a failure. The criterion passes on AMBER (9/9) and on Object HalBench. With the
COCO-fitted probe and no AMBER label, however, the AMBER strong baseline \emph{fails} its criterion (6/9):
on a new image distribution the strong baseline needs a small amount of in-domain fitting.
SENTINEL passes on its own (S2: $+1.86$ $[+1.07,+2.64]$); its full recall is far below the base
model's (57.0 vs.\ 79.7), which matters for comparisons against the truncated base model
(Section~\ref{sec:limits}).

\textbf{Newer backbones.} On Qwen2.5-VL-7B the base model's $\Gp$ is $+4.46$ $[+3.43,+5.36]$ (P4, PASS);
seven of our eight Qwen checkpoints have intervals above zero (secondary); SENTINEL on Qwen has
$+4.83$ $[+3.81,+5.80]$ (S4, PASS, full recall 67.2 vs.\ 69.0). On InternVL2.5-8B all eight of our
checkpoints have intervals above zero ($\Gp$ from $+0.92$ to $+3.89$; primary, PASS), and the base model's
$\Gp$ is $+2.94$ $[+1.85,+3.95]$ (reported). A probe fitted on LLaVA-1.5 and applied to Qwen2.5-VL
unchanged still beats truncation ($+2.08$ $[+1.09,+2.96]$; secondary).

\textbf{Operating points.} In a post-hoc analysis the strong baseline meets its bar at every operating point
from 70\% to 90\% of full recall: at least 8 of 9 LLaVA-1.5 systems on COCO (CHAIR-DPO's interval contains
zero from 75\% to 90\%), 9 of 9 on AMBER, and at least 7 of 8 InternVL2.5 checkpoints. The pre-specified
point remains 80\%. Applied to the untuned base model and read at each trained method's own recall, the
probe matches or beats 4 of the 8 methods (HA-DPO, OPA-DPO, EOS, CSR-iter3; secondary).

\textbf{Realized as text.} Everything above scores object sets, but a method has to emit a caption. We
therefore delete whole sentences: each sentence is scored by the highest probe score among the objects it
introduces, sentences are removed under a global threshold, and the remaining string is re-scored by the
evaluator, against keeping the first $k$ sentences. On Qwen2.5-VL this co-primary criterion (P6; added after
an exploratory LLaVA analysis and before any Qwen caption was scored) passes at $+5.45$ $[+4.38,+6.45]$. On
LLaVA-1.5, in that exploratory analysis, sentence deletion beats keeping the first $k$ sentences on all six
systems then evaluated ($+0.75$ to $+1.92$ points), about half the object-level gain. Qwen packs fewer
objects per sentence, so deleting a sentence costs fewer correct objects.

\textbf{Negative controls (exploratory).} Pruning by the COCO priors alone or by the model's own confidence
alone is \emph{worse} than truncation on all six LLaVA-1.5 systems evaluated (mean $\Gp$ $-2.20$ and $-2.38$,
0/6), whereas the surface features alone and the probe without priors both win 6/6. Truncation is itself a
perfect position-based pruner, so position-free variants start at a disadvantage; the result means that
these signals add nothing over saying less, not that they are uninformative.

\section{Cheap labels for preference training}
\label{sec:labels}
If cheap signals account for most of what preference training removes, can they also supply the
preference labels? We train DPO on the model's own samples, labelled in different ways, and price each
labelling source with the same matched-recall measure.

\subsection{Measure, labellers and recipe}

For a checkpoint $c$, $G(c)$ is the hallucination rate ($\chd$, held-out $B{+}C$) of the base model's
captions truncated to $c$'s own full recall, minus $c$'s own rate: its gain beyond saying less. For a
labelling source $X$ trained exactly like a ground-truth arm, the \textbf{recovery ratio}
$R_X = \overline{G}(X) / \overline{G}(\mathrm{GT})$ is the share of the GT-labelled gain it recovers. A
labelling source can also help by making the model terser in a way the recall match does not remove,
so each comparison is checked against a \textbf{matched random} arm that, image by image, removes as much
text as the probe arm does but chooses what to remove at random.

\textbf{Labellers.} On four samples per training image (COCO \texttt{train2014}), objects are flagged as
hallucinated by: the COCO annotation (GT); the base model's frozen probe (Section~\ref{sec:probe}); or an
object-level CLIP score, $s(c) = \cos(\text{image}, \text{``a photo of a }c\text{''}) - \frac{1}{80}\sum_k
\cos(\text{image}, \text{``a photo of a }k\text{''})$ over the 80 COCO classes, which removes an image's overall
bias towards any text. The probe and CLIP thresholds are set by the same rule on batch $A$ only (the
flagged fraction equals batch $A$'s true hallucination rate); neither looks at any annotation of the
training images. The CLIP labeller is an object-level adaptation inspired by CLIP-DPO \cite{clipdpo}, which ranks
sampled responses as wholes.

\textbf{Pairs and recipe.} At response granularity the chosen sample has the fewest flagged objects and the
rejected one the most; at object granularity the rejected sample is paired with itself minus the
sentences containing flagged objects. Each arm has 2527 pairs, two granularities and two seeds, and the
same LoRA-DPO recipe ($r$ 16, $\alpha$ 32, $\beta = 0.1$, learning rate $2{\times}10^{-5}$, 2 epochs, with an
SFT term of weight 1).

\subsection{Results}

\begin{table}[t]
\centering
\footnotesize
\setlength{\tabcolsep}{2pt}
\begin{tabular}{llccc}
\toprule
backbone & labels & mean $G$ & recovery $R$ & vs.\ matched random \\
\midrule
Qwen2.5-VL-7B & GT        & $+3.81$ & 1 & \\
              & probe     & $+2.67$ & $0.702$ $[0.560,0.849]$ (P1 PASS) & $+2.21$ $[+1.57,+2.87]$ (P2$'$ PASS) \\
              & CLIP      & $+3.31$ & $0.869$ $[0.736,1.013]$ (C2 PASS) & $+2.85$ $[+2.12,+3.59]$ (C3, secondary) \\
              & matched random & $+0.46$ & & \\
\multicolumn{5}{l}{\quad probe $-$ CLIP: $-0.64$ $[-1.20,-0.10]$ (C1, two-sided): CLIP labels buy more} \\
\midrule
InternVL2.5-8B & GT       & $+3.67$ & 1 & \\
               & probe    & $+2.68$ & $0.730$ $[0.554,0.914]$ (PB1 PASS) & $+1.29$ $[+0.52,+2.02]$ (PB2 PASS) \\
               & matched random & $+1.39$ & & \\
\bottomrule
\end{tabular}
\caption{\textbf{Cheap labels for DPO.} $G$: gain beyond saying less on COCO held-out images (points of
$\chd$); means over 4 checkpoints per arm on Qwen2.5-VL (2 granularities $\times$ 2 seeds) and 2 on
InternVL2.5 (response granularity, 2 seeds; 1492 common images). Recovery $R = \overline{G}(X)/\overline{G}(\mathrm{GT})$
with the pre-set bar $R \ge 0.667$ and lower bound $> 0.333$; intervals: 1000-draw image bootstrap.}
\label{tab:labels}
\end{table}

\textbf{Probe labels.} On Qwen2.5-VL, DPO on probe labels recovers $0.70$ of the GT-labelled gain
(Table~\ref{tab:labels}; P1 PASS). The recovery is not stable across seeds: within seed 0 it is 0.589 and
within seed 1 0.825, so the secondary per-seed criterion fails and we report only the pooled ratio. On
InternVL2.5 the probe arm recovers $0.73$ (PB1 PASS), with per-seed ratios 0.684 and 0.778 (secondary, PASS;
seed 0 close to the bar).

\textbf{Against random labels.} The originally specified comparison with unmatched random labels (P2) is
\emph{not evaluable}: all four random-label checkpoints failed the pre-set validation gate (their
hallucination rate on validation images dropped by less than 10\%) and were excluded as not trained,
leaving that arm empty. The frozen script's mechanical reading for a non-passing P2 (``the gain comes from
the recipe, not the labels'') therefore rests on an empty comparison. That random arm also removed about
twice as many sentences as the probe arm, which would have confounded the comparison. We then specified,
and froze before training it, a matched-random arm in which the validation gate is reported but excludes
nothing. Against it the probe arm wins on Qwen2.5-VL ($+2.21$ $[+1.57,+2.87]$, P2$'$ PASS, also within each
seed) and on InternVL2.5 ($+1.29$ $[+0.52,+2.02]$, PB2 PASS). Matched random labels recover 0.38 of the GT
gain on InternVL2.5 (reported) and give a mean $G$ of only $+0.46$ on Qwen2.5-VL.

\textbf{CLIP labels.} On Qwen2.5-VL, CLIP labels recover $0.87$ (C2 PASS) and beat matched random labels by
$+2.85$ (secondary). The two-sided comparison with probe labels, whose three possible readings were fixed
in advance, comes out against our own probe: CLIP labels buy more beyond saying less
($-0.64$ $[-1.20,-0.10]$, C1), with the same sign within each seed ($-1.00$ and $-0.28$). The CLIP arm was
run on Qwen2.5-VL only.

\textbf{AMBER (descriptive).} On AMBER the probe arm's recovery on Qwen2.5-VL is $0.808$ $[0.579,1.041]$ and
its advantage over matched random $+1.17$ $[+0.69,+1.66]$; on InternVL2.5 the advantage over matched random
($+0.90$ $[+0.36,+1.49]$) carries over but the recovery ratio is not evaluable (the GT arm's mean gain is not
reliably positive on AMBER). CLIP and probe labels cannot be distinguished on AMBER ($-0.33$
$[-0.81,+0.13]$), with CLIP recovering $0.971$ $[0.767,1.227]$ and beating matched random by $+1.49$
$[+0.98,+2.03]$. The first run of this CLIP analysis found that the CLIP checkpoints' AMBER labels had not
been generated (an omission in our pipeline); after generating them, the unchanged script was re-run, and
both outputs are kept.

\textbf{A post-hoc observation.} On the same 153{,}225 training-sample rows, the probe ranks hallucinations
better than CLIP (AUROC 0.841 vs.\ 0.813), with similar flag precision (0.471 vs.\ 0.488); CLIP flags fewer
objects (flag rate 0.148 vs.\ 0.194; recall 0.436 vs.\ 0.552). Yet the CLIP-labelled models are better. A
labeller's AUROC therefore does not by itself predict its value as a training signal; this is one
comparison, and we offer no explanation.

Cheap object-level labels calibrated on 500 annotated images thus recover most of what ground-truth labels
buy beyond saying less, on two backbones for the probe and on one for CLIP, and the recovery ratio against a
matched random arm is a direct way to compare labelling sources.

\section{Human evaluation}
\label{sec:human}
All results so far depend on an automatic evaluator. We asked people three questions, with criteria and an
agreement gate fixed in advance.

\textbf{Design.} Task A (600 items) checks the evaluator: is an object that CHAIR calls hallucinated (or real)
in the image? Task B (120 images) checks the deletion: is the probe-edited caption still fluent, and is each
deleted or kept sentence wrong? Task C (210 pairs, seven trained methods) is a blind matched-recall
comparison: is the base-plus-probe caption more wrong than the trained method's caption at the same recall?
An LLM pre-labelled every item. Two annotators labelled an independent subset (60 / 15 / 25 items for
A / B / C) without seeing those labels, and reviewed the remaining items starting from the LLM's labels,
changing 28.7\% / 51.4\% / 29.2\% of them. The annotators saw the English captions together with an
LLM-produced Chinese reference translation (a pre-specified amendment). A task's criteria were to be judged
only if the mean LLM--human agreement on the independent subset reached Cohen's $\kappa \ge 0.40$; otherwise
they are reported as descriptive only.

\textbf{Agreement.} Only task B passed the gate (mean LLM--human $\kappa$ 0.523; human--human 0.549). Task A
(0.296; human--human 0.055) and task C (0.351; 0.284) did not, so the two \emph{primary} criteria, HA (task A)
and HC (task C), are descriptive only. The second annotator later reported problems with their own task-A
answers; in their task-B review two fields were left empty and were filled with the LLM's original answers.

\textbf{Deletion quality (task B, secondary criteria, both PASS).} Probe-guided sentence deletion keeps
captions readable (mean fluency 4.77 before, 4.62 after; the bar allowed a drop of 0.5), and the sentences
it deletes are judged wrong far more often than those it keeps (0.535 vs.\ 0.243; difference $+0.292$
$[+0.204,+0.374]$; HB). They are also judged wrong more often than the sentences truncation to the same
recall would delete (0.535 vs.\ 0.451; $+0.083$ $[+0.030,+0.139]$; HB-iii).

\textbf{The evaluator (task A, primary, descriptive only).} Of the objects CHAIR calls hallucinated, the
fraction that people judge absent from the image is 0.531 $[0.434,0.628]$ on COCO and 0.581 $[0.442,0.721]$
on AMBER, below the pre-set bar of 0.80. After seeing these results, a third annotator independently redid all
600 task-A items (an additional, post-hoc report; the frozen verdicts are unchanged). Their agreement with the
LLM passes the gate ($\kappa$ 0.566); under their labels the precision is 0.683 $[0.603,0.762]$ on COCO and
0.707 $[0.561,0.829]$ on AMBER, still below 0.80. Their instructions contained one additional sentence: judge
only whether the highlighted object is in the image, not whether the whole sentence is right. On the
independent subset the third annotator agrees with the first annotator ($\kappa$ 0.649) but not with the second
(0.061).

\textbf{Matched-recall blind comparison (task C, primary, descriptive only).} For 4 of the 7 methods (CHAIR-DPO,
HA-DPO, OPA-DPO, POVID), people judge the base-plus-probe caption not more wrong than the trained method's in at
least half the pairs with a Wilson lower bound above 0.40; the criterion required 6. RLAIF-V (0.464), CSR-iter3
(0.536) and EOS (0.552) do not reach it.

Human judgement thus supports the deletion step --- the probe removes wrong sentences, more precisely than
truncation, without breaking the caption --- but it does not certify the evaluator, whose estimated precision
is below our bar under every set of human labels we have (Section~\ref{sec:limits}).

\section{The rest}
\label{sec:rest}
\textbf{RLAIF-V.} It has the lowest share on COCO (0.59), AMBER (0.83) and Object HalBench (0.62), and is below
the bar on COCO and OHB; in the blind comparison it is the method for which people least often judge the
base-plus-probe caption not more wrong (0.464). An exploratory COCO analysis suggested a specific reading ---
that RLAIF-V learns information the cheap signals lack --- and its pre-specified replication on AMBER failed.
We therefore state only that RLAIF-V is the method the cheap account explains least.

\textbf{No discriminative perception gained (POPE).} If what training buys is mostly cheap pruning, it should
not improve the model's answers to yes/no questions about object presence. The primary POPE criterion asked
whether the methods' accounted shares predict their POPE gains (Spearman $\rho$ over the 8 methods, expected
negative); it failed ($\rho = +0.443$, one-sided $p = 0.871$), so POPE is descriptive only. On the adversarial
split, only 1 of the 8 methods improves AUROC with an interval above zero (HALVA, $+0.0007$
$[+0.0002,+0.0012]$); RLAIF-V ($-0.0073$) and HA-DPO ($-0.0052$) are significantly worse than the base model
(secondary).

\textbf{On-policy data do not explain the remainder.} The primary hypothesis of our controlled Qwen2.5-VL
experiment was that training on the model's own samples makes its deletions harder for the probe to
reproduce (lower share). The opposite holds: on-policy checkpoints have the \emph{higher} share
($D_{\mathrm{on}} = +0.189$ $[+0.099,+0.266]$, both seeds positive), so the hypothesis is not supported.

\section{Limitations}
\label{sec:limits}
\begin{itemize}
\item \textbf{Evaluator dependence.} Every number rests on an automatic object evaluator (CHAIR's synonym
lists on COCO, the official evaluator on AMBER). People judge only 0.53--0.58 of its hallucination calls to be
correct (0.68--0.71 under a third annotator), below our pre-set bar of 0.80, and human--human agreement on that
task was low; this is in line with Feng et al.~\cite{feng2024}. COCO's annotation also misses objects: of the
objects that the COCO annotation calls hallucinated in a sample from our training images, people judge 0.644
to be present ($n = 45$).
A post-hoc check found no evidence that the evaluator's precision differs between the base model and the
trained methods (difference $+0.019$ $[-0.118,+0.193]$), but it can only exclude differences larger than about
15 points. System-dependent evaluator errors would affect comparisons between a method and the truncated base
model (the accounting and the gains $G$ of Section~\ref{sec:labels}); the strong baseline compares two edits of
the same system's captions.
\item \textbf{The probe inherits COCO's labels.} It is fitted on COCO annotations, and on the items where CHAIR
calls an object hallucinated its score separates the human ``absent'' and ``present'' judgements only weakly
(AUROC 0.579 $[0.470,0.682]$; post hoc).
\item \textbf{Low recall.} Comparisons of a system with its base model truncated to the system's own recall were
found, in exploration, to be unreliable when that recall is far below the base model's. SENTINEL on LLaVA-1.5
(57.0 vs.\ 79.7) is in that region; the checkpoints of Section~\ref{sec:labels} range from 56.7 to 68.5 against
the base model's 69.0.
\item \textbf{Where the accounting fails.} On InternVL2.5 the accounting fails on COCO. On the newer backbones,
the accounting rests on our own checkpoints and on one public method (SENTINEL); we do not generalize to public
methods on new backbones.
\item \textbf{Training-data overlap.} Of the 1500 held-out images, 87.8\% belong to COCO \texttt{train2017} and
41.9\% occur in LLaVA-1.5's instruction data. All comparisons are made within the same image and caption, but
split by that overlap the strong baseline reaches 7 of 9 systems in each half, below the 8 of 9 our post-hoc
rule required, so we do not claim that it holds on images the models have never seen; the non-COCO check is
AMBER. The training data of Qwen2.5-VL and InternVL2.5 cannot be checked.
\item \textbf{Scope.} We study object existence in detailed captions; attributes and relations are not
evaluated. The probe is a post-hoc reranker: two of its strongest features need the finished caption. It needs
a small amount of in-domain labelled data (the zero-shot AMBER strong baseline failed). Sentence deletion is a
crude edit that does not repair references across sentences.
\item \textbf{Sample sizes.} The independent human subsets are small (60 / 15 / 25 items); with only their labels
every human criterion fails, with wide intervals. Self-trained arms have two to four checkpoints.
\end{itemize}

\section{Conclusion}
Measured against the same model saying less, most of what current anti-hallucination preference training
removes on detailed captioning is reproducible by a frozen eleven-feature probe over cheap post-hoc signals,
fitted on 500 images: for seven of eight public LLaVA-1.5 methods on COCO, all eight on AMBER, a later public
method on two backbones, and our own checkpoints on Qwen2.5-VL --- but not for our InternVL2.5 checkpoints on
COCO, and least of all for RLAIF-V. Base model plus probe beats saying less on every backbone we tested, and
the same cheap signals, used as preference labels, recover most of what ground-truth labels buy. We suggest
that claimed reductions in hallucination be reported against a recall-matched control and a cheap post-hoc
baseline, and that labelling sources be priced by what they recover against matched random labels.

\appendix
\section{All pre-specified primary criteria and outcomes}
\label{app:prereg}
Table~\ref{tab:prereg} lists every primary criterion of the paper with its outcome: 32 criteria, of which
23 pass, 5 fail, 1 is not evaluable, 2 were turned into descriptive-only analyses by the agreement gate, and
1 is a two-sided comparison whose fixed reading is unfavourable to our probe. We apply no family-wise
correction; instead every claim in the main text rests on a primary criterion stated with its bar, secondary
and descriptive analyses are labelled as such, and every primary outcome is reported. The list was produced
by a script that reads each criterion's role from the frozen analysis code and its outcome from the frozen
output, and matched our hand count. Superseded outputs (the first AMBER version, which had a label-leakage
bug) are not counted.

\textbf{Documentation of the freeze.} A: the md5 freeze of the script and its input manifest predates the
outputs it judges. B: the order is documented only by dated statements in the scripts themselves (the first
LLaVA and Qwen analyses). C: re-frozen once after a bug was found in the first version, both versions
reported (AMBER). P6 was added after an exploratory LLaVA analysis and before any Qwen caption was scored.

\begin{table}[!ht]
\centering
\scriptsize
\setlength{\tabcolsep}{3pt}
\newcolumntype{L}[1]{>{\raggedright\arraybackslash}p{#1}}
\begin{tabular}{L{3.4cm}L{3.0cm}L{4.2cm}L{2.9cm}c}
\toprule
analysis & criterion & bar & outcome & doc. \\
\midrule
LLaVA, COCO, 5 methods & P1 accounting & share $\ge 0.667$ for $\ge 4/5$ & PASS (4/5) & B \\
                       & P2 strong baseline & $\Gp$ CI $>0$ for $\ge 5/6$ & PASS (6/6) & B \\
Qwen2.5-VL base        & P4 strong baseline & $\Gp$ CI $>0$ & PASS ($+4.46$) & B \\
                       & P6 sentence deletion & $\Gs$ CI $>0$ & PASS ($+5.45$) & B \\
LLaVA, 3 added methods, batch $C$ & E1 accounting & share $\ge 0.667$ for $\ge 2/3$ & PASS (3/3) & B \\
                       & E2 strong baseline & CI $>0$ for $\ge 2/3$ & PASS (3/3) & B \\
LLaVA, COCO, 8 methods & Q1 accounting & share $\ge 0.667$ for $\ge 6/8$ & PASS (7/8) & B \\
                       & Q2 strong baseline & CI $>0$ for $\ge 8/9$ & PASS (8/9) & B \\
AMBER                  & Q1 accounting & $\ge 6/8$ & PASS (8/8) & C \\
                       & Q2 strong baseline & $\ge 8/9$ & PASS (9/9) & C \\
AMBER, RLAIF-V reading & R1 & highest; paired CI $>0$ vs.\ $\ge 6/7$ & FAIL & C \\
AMBER zero-shot        & Q1 accounting & $\ge 6/8$ & PASS (6/8) & A \\
                       & Q2 strong baseline & $\ge 8/9$ & \textbf{FAIL} (6/9) & A \\
Object HalBench        & Q1 accounting & $\ge 6/8$ & PASS (6/8) & A \\
                       & Q2 strong baseline & $\Gp>0$ for $\ge 8/9$; mean CI $>0$ & PASS & A \\
POPE                   & P1 share vs.\ $\Delta$AUROC & Spearman $\rho < 0$, one-sided & \textbf{FAIL} ($\rho = +0.443$) & A \\
Qwen, 2$\times$2 controlled & H1 on-policy & $D_{\mathrm{on}}$ CI $<0$; seeds $<0$ & \textbf{FAIL} ($+0.189$) & A \\
Probe-label DPO, Qwen  & P1 recovery & $R \ge 0.667$, CI low $> 0.333$ & PASS (0.702) & A \\
                       & P2 vs.\ random labels & CI $>0$ & not evaluable & A \\
Matched-random follow-up & P2$'$ vs.\ matched random & CI $>0$ & PASS ($+2.21$) & A \\
InternVL2.5, own DPO   & Q1 accounting & $\ge 3/4$ designs & \textbf{FAIL} (2/4) & A \\
                       & Q2 strong baseline & $\ge 6/8$ checkpoints & PASS (8/8) & A \\
Probe-label DPO, InternVL & PB1 recovery & $R \ge 0.667$, CI low $> 0.333$ & PASS (0.730) & A \\
                       & PB2 vs.\ matched random & CI $>0$ & PASS ($+1.29$) & A \\
Human evaluation       & HA evaluator precision & $\ge 0.80$ on COCO and AMBER & descriptive only & A \\
                       & HC blind comparison & $\ge 6/7$ methods & descriptive only & A \\
CLIP-label DPO, Qwen   & C1 probe $-$ CLIP & two-sided, 3 fixed readings & CLIP buys more ($-0.64$) & A \\
                       & C2 recovery & $R \ge 0.667$, CI low $> 0.333$ & PASS (0.869) & A \\
SENTINEL               & S1 LLaVA accounting & share $\ge 0.667$ & PASS (0.82) & A \\
                       & S2 LLaVA strong baseline & CI $>0$ & PASS ($+1.86$) & A \\
                       & S3 Qwen accounting & share $\ge 0.667$ ($\ge 30$ objects) & PASS (0.99) & A \\
                       & S4 Qwen strong baseline & CI $>0$ & PASS ($+4.83$) & A \\
\bottomrule
\end{tabular}
\caption{\textbf{All 32 primary criteria.} doc.: how the freeze is documented (see text). E0, a calibration gate
on batch $C$, decided whether E1 and E2 could be read and passed. The 5-method and 8-method LLaVA analyses overlap (the latter
extends the former to three added methods on the same images); both are listed because both were
pre-specified.}
\label{tab:prereg}
\end{table}

\end{document}